\def\finalversion{}
\def\arxivversion{}
\documentclass{article}
\PassOptionsToPackage{sort&compress}{natbib}
\usepackage[table]{xcolor} %
\ifdefined\finalversion
  \usepackage[final]{corl_2026}
\else
  \usepackage{corl_2026}
\fi
\usepackage[T1]{fontenc}
\usepackage{amsmath,amssymb,bm}
\usepackage{booktabs,multirow,array}
\usepackage{graphicx}
\usepackage{adjustbox}
\usepackage{enumitem}
\usepackage{microtype}
\usepackage{placeins}
\usepackage[font=small,labelfont=bf,skip=3pt]{caption}
\setlist{nosep,leftmargin=1.3em}

\makeatletter
\ifdefined\arxivversion
  \renewcommand{\@noticestring}{Preprint.}
\else\ifdefined\finalversion
  \renewcommand{\@noticestring}{Workshop paper at the 10th Conference on Robot Learning (CoRL 2026), Austin TX, USA.}
\else
  \renewcommand{\@noticestring}{Submitted to a workshop at the 10th Conference on Robot Learning (CoRL 2026). Do not distribute.}
\fi\fi
\makeatother

\newcommand{\R}{\mathbb{R}}
\newcommand{\E}{\mathbb{E}}
\newcommand{\pz}{$\pi_0$}
\newcommand{\ctrl}[2]{Ctrl(#1,\,#2\,s)}

\title{Embodiment-Classifier Guidance: Transferring Frozen VLAs Across Robot Dynamics without Demonstrations}

\author{
  Devin Han\\
  UC Berkeley\\
  \texttt{devinhan@berkeley.edu}
  \And
  Hershraj Niranjani\\
  UC Berkeley\\
  \texttt{hershraj@berkeley.edu}
}

\hypersetup{pdftitle={Embodiment-Classifier Guidance: Transferring Frozen VLAs Across Robot Dynamics without Demonstrations},
            pdfsubject={Workshop paper, 10th Conference on Robot Learning (CoRL 2026)}}
\ifdefined\finalversion
  \hypersetup{pdfauthor={Devin Han, Hershraj Niranjani}}
\fi
\ifdefined\arxivversion
  \hypersetup{pdfsubject={Preprint}}
\fi

\begin{document}
\maketitle

\begin{abstract}
Flow-based policies such as \pz{} are trained on a fixed set of embodiments, and transferring them to new ones is an open problem.
To adapt a frozen \pz{} to a new embodiment without additional demonstrations or parameter updates, we train a lightweight embodiment classifier to distinguish successful action chunks on the source body from those on the target body, using a few calibration tasks.
On transfer tasks that \pz{} saw on its source body but the classifier was never trained on, we nudge each denoising step of \pz{} with the classifier's gradient, which estimates the direction from the distribution of successful actions on the source body toward that on the target body.
On the LIBERO simulation benchmark with four modified controllers, \emph{embodiment-classifier guidance} raises \pz{}'s average success on the target bodies from 72\% to 80\% across 12 suite--controller pairs, with a 31\% relative reduction in failures (11 of 12 gain; 5 significant after Holm correction).
In simulation, guidance is data-efficient, needing few target-body rollouts per task, which makes it promising for sim-to-real transfer, where rollouts are costly.
Guidance is mixed on weak wrists and harmful on a locked wrist, when the classifier lacks data or favors infeasible motions.
Collecting more target-body data with guided \pz{} can fill such gaps: on one weak-wrist arm, guided \pz{} solved a task on which unguided \pz{} had zero successes in the original data, and retraining the classifier on the guided successes raised success from 52.5\% to 70.7\%.
\end{abstract}


\begin{figure}[!h]
\vspace{-12pt}
\centering
\includegraphics[width=\linewidth]{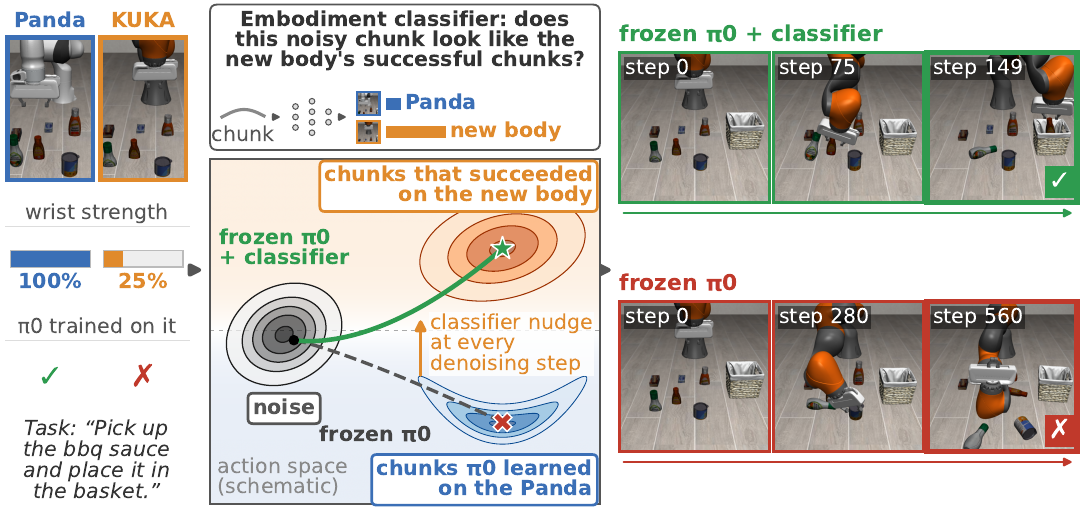}
\caption{\textbf{Embodiment-classifier guidance steers a frozen \pz{} toward plans that succeed on a target body.} \pz{}, trained on a Panda, runs on a KUKA arm with 25\% wrist strength. Middle (schematic): the classifier nudges every denoising step. Right: a selected scene where guidance succeeds and unguided \pz{} fails.}
\label{fig:teaser}
\end{figure}

\section{Introduction}
\label{sec:intro}
Flow-based VLA policies such as \pz{}~\cite{black2024pi0} are trained on a fixed set of robot \emph{bodies} (an arm together with its low-level controller), and deploying them on new ones remains an open challenge.
On a body with different dynamics (a softer controller, command latency, a weaker wrist motor, another arm), the same policy can fail even when the scene and the task are unchanged (Fig.~\ref{fig:teaser}).
Such gaps separate simulation from the lab and the lab from the field; closing them usually takes target-body demonstrations and fine-tuning~\cite{hu2022lora} or a per-task steering policy learned from reward~\cite{wagenmaker2025dsrl}.

\textbf{Background.} \pz{} maps an observation (camera images, robot state and a language instruction) to an \emph{action chunk}: the next 50 actions of 7 numbers each, which together form a single point in a 350-dimensional space.
Like an image diffusion model, it generates this point by sampling: starting from Gaussian noise, a learned \emph{velocity} field moves it toward a plausible chunk over 10 denoising steps (Fig.~\ref{fig:teaser}, middle).
The robot executes only the first 5 actions of each chunk, discards the rest, and samples a new chunk from the next observation.

\textbf{Approach.} We instead change only how a frozen flow policy samples.
\pz{} has no input that says which body it controls, so it samples chunks that suit its source body.
We train a small \emph{embodiment classifier} to tell \pz{}'s successful action chunks on the source body from those on the target body, and add its gradient to \pz{}'s velocity at every denoising step.
Intuitively, the classifier scores how much a noisy chunk looks like a target-body success, so its gradient points toward chunks that work on the target body, while \pz{}'s velocity field still provides the task-level plan, such as which object to reach and where to place it.
As we derive in Eq.~\eqref{eq:ratio}, with an exact classifier this gradient, suitably scaled, is exactly the velocity difference between the two bodies' distributions of successful chunks; under the assumptions in App.~\ref{app:theory}, guided sampling would then draw from the target body's successes, and we hypothesize that these assumptions hold approximately in practice.

\textbf{Relation to prior work.} Similar classifier guidance has been used to refine image diffusion models~\cite{kim2023discriminator} and to adapt pretrained diffusion models to data-scarce target domains~\cite{ouyang2024tgdp}; we apply it to robot embodiment transfer, with the body as the classifier's label. Unlike embodiment-aware guidance from a known controller's tracking cost~\cite{gupta2025umionair}, our guidance is learned from successful rollouts.
Because the classifier sees no images, instruction or task label, only noisy action chunks (which implicitly carry the end-effector pose; Sec.~\ref{sec:method}), guidance with a classifier trained on a few calibration tasks can improve \pz{}'s target-body success on other \emph{transfer tasks}, which the classifier was never trained on. \pz{} itself was trained on these tasks with its source body, so our tests measure transfer across robot dynamics, not generalization to new tasks.

LIBERO~\cite{liu2023libero} is a simulated benchmark of tabletop manipulation tasks in four suites. Our contributions, evaluated on it with one \pz{} checkpoint and paired, pre-registered tests, are:
\begin{itemize}
\item \textbf{Embodiment-classifier guidance:} a way to steer a frozen flow VLA to a new body with a classifier learned from successful rollouts, with no demonstrations or fine-tuning (Sec.~\ref{sec:method}).
\item \textbf{Guidance improves frozen \pz{} on most controller changes:} with four modified controllers, it raises average success from 72\% to 80\% over 12 suite--body pairs, with a 31\% relative reduction in failures (gains in 11 of 12; Sec.~\ref{sec:res-ctrl}).
\item \textbf{Embodiment-classifier guidance is data-efficient in simulation:} the classifier trains on the successful episodes among about 23 target-body rollouts per calibration task on LIBERO-Object (pick-and-place), with no demonstrations, reward model or updates to \pz{}. In a preliminary check on one body, an untuned LoRA fine-tune with the same collection budget lowered success (Sec.~\ref{sec:res-grad}).
\item \textbf{It is not a universal fix:} it is mixed on weak-wrist arms and hurts where the body cannot execute \pz{}'s plans (Sec.~\ref{sec:res-mixed}).
\item \textbf{Guided rollouts fill data gaps:} collecting more target-body data with guided \pz{} finds successes on a task where unguided \pz{} had none, and retraining the classifier on the guided successes lifts a weak-wrist KUKA arm from 52.5\% to 70.7\%, while the same amount of extra data from unguided \pz{} reaches only 57.8\% (Sec.~\ref{sec:res-gsc}).
\end{itemize}

\section{Embodiment-Classifier Guidance}
\label{sec:method}
\textbf{Setting.}
\pz{} maps an observation $o$ (two camera images, robot state, instruction) to an action chunk $a\in\R^{50\times7}$.
It is a flow-matching model~\cite{lipman2023flow}: with the noised chunk $x_t=t\,\varepsilon+(1-t)\,a$ and flow time $t\in[0,1]$, a network $v(x_t,t,o)$ predicts the velocity $\varepsilon-a$, and sampling takes 10 Euler steps from $x_1=\varepsilon$,
\begin{equation}
x_{t-0.1}=x_t-0.1\,v(x_t,t,o),\qquad t=1,0.9,\dots,0.1,
\label{eq:sample}
\end{equation}
after which the robot executes 5 actions and \pz{} plans again.
\pz{} was trained on the \emph{source body} (A); we deploy it on a \emph{target body} (B), where \emph{unguided \pz{}} is the baseline.
We may add a term to \pz{}'s velocity but never change its weights.
On each body, we may run \pz{} on a few \emph{calibration tasks} and observe which episodes succeed (2,500 \pz{} queries per task, about 23 episodes on LIBERO-Object); \emph{transfer tasks} get no demonstrations, reward or target-body data.

\textbf{Guidance as a density ratio.}
Let $D_A$ and $D_B$ hold chunks from successful episodes of unguided \pz{} on each body's calibration tasks, and let $q^X_t$ be the distribution of $D_X$'s chunks noised along \pz{}'s path.
On this path $v(x,t)=-\big(x+t\,\nabla_x\log p_t(x)\big)/(1-t)$ (App.~\ref{app:theory}), so two chunk distributions differ in velocity by exactly
\begin{equation}
v^B(x,t)-v^A(x,t)=-\frac{t}{1-t}\,\nabla_x\log\frac{q^B_t(x)}{q^A_t(x)} .
\label{eq:ratio}
\end{equation}
A small noise-conditioned transformer $\ell_\phi(x_t,t)$ (0.8M weights), trained with the logistic loss to tell $D_B$ from $D_A$, estimates this log ratio: its optimum is $\ell^\star=\log q^B_t/q^A_t$.

\textbf{Guided sampler.}
We replace the log ratio with $\ell_\phi$ and add a weight $w$, a clamp and a cap:
\begin{equation}
\delta v=-\,w\,\frac{t}{\max(1-t,\,0.05)}\,\nabla_x\ell_\phi(x,t),\qquad
\hat v = v + \delta v\cdot\min\!\Big(1,\;\frac{0.2\,\lVert v\rVert}{\lVert\delta v\rVert}\Big).
\label{eq:guide}
\end{equation}
The clamp keeps the factor finite at $t=1$, the cap limits each nudge to 20\% of \pz{}'s step, and $w\in\{1,2,4\}$ is picked on the calibration tasks; these, and a classifier that does not see $o$ directly, are heuristics (App.~\ref{app:theory}).

\textbf{What the classifier learns.}
\pz{} nearly always succeeds on its source body, so $D_A$ is roughly \pz{}'s own chunks and $D_B$ is those chunks filtered by success on the target body: the aim is \pz{} reweighted by the chance that a chunk succeeds there (App.~\ref{app:theory}).
The classifier sees no images, instruction or task label, but \pz{}'s action space subtracts the current end-effector pose (App.~\ref{app:setup}), so the pose is implicitly visible.

\textbf{Guided self-collection.}
If guided \pz{} solves calibration tasks where $D_B$ has few successes, we add the successes of one round of guided rollouts to $D_B$ and retrain the classifier.

\section{Experimental Setup}
\label{sec:setup}
\textbf{Policy and tasks.}
We use the frozen public \pz{} LIBERO checkpoint~\cite{black2024pi0} in LIBERO~\cite{liu2023libero} (robosuite~\cite{zhu2020robosuite}), a simulated benchmark of tabletop manipulation tasks in four suites that vary the objects (Object), spatial layouts (Spatial) and goals (Goal), plus long-horizon tasks (LIBERO-10).
Each suite has four calibration and four transfer tasks (App.~\ref{app:setup}), all of which \pz{} saw on its source body.
Episode limits are about twice LIBERO's (capped at 980 steps for LIBERO-10), so slow bodies are not failed for speed alone, and every condition sees the same 200 transfer-task episodes.

\textbf{Target bodies} (Table~\ref{tab:bodies}) form three families.
\emph{Controller changes} \ctrl{$k$}{$d$} are the Panda with its controller stiffness scaled by $k$ and its commands delayed by $d$ seconds: \ctrl{1}{0.2}, \ctrl{0.5}{0.2}, \ctrl{0.5}{0.3}, \ctrl{0.33}{0.3} and six sweep bodies.
\emph{Weak wrists} are the KUKA, Kinova and Sawyer arms (with the Panda gripper) and the Panda, with wrist motors at 25\%, 10\%, 25\% and 10\% strength.
\emph{Other faults}: a UR5e capped at 0.3\,rad/s, a 5$\times$ slower gripper, a worn Panda, a locked wrist.

\textbf{Statistics.}
We report success rates over 200 transfer-task episodes and paired differences in percentage points (points) with 95\% paired-bootstrap confidence intervals (CIs). Conditions, seeds and criteria (a gain of at least +5 or +10 points with a CI above zero) were committed before data collection (exceptions: App.~\ref{app:stats}).

\section{Results}
\label{sec:results}

\begin{figure}[t]
\centering
\includegraphics[width=\linewidth]{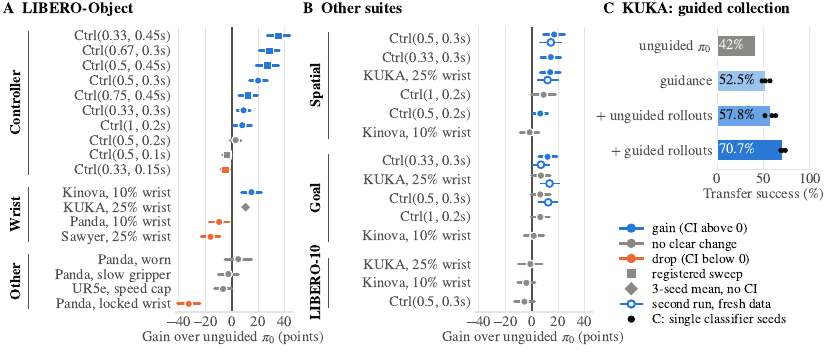}
\caption{\textbf{Transfer-task gain over unguided \pz{}} (percentage points; 95\% paired-bootstrap CIs). \textbf{A:} LIBERO-Object by body family (squares: registered sweep, where \ctrl{0.75}{0.45} is below the registered floor and reported only; diamond: 3-seed mean). \textbf{B:} other suites (hollow: second runs, i.e., full repeats with new target-body data). \textbf{C:} KUKA arm (25\% wrist), classifier retrained with extra successes from unguided or guided \pz{} with the same number of extra target-body queries (bars: 3-seed means; dots: seeds).}
\label{fig:main}
\end{figure}

\subsection{Controller changes: guidance helps}
\label{sec:res-ctrl}
Across the four suites, guidance improves 11 of the 12 suite--body pairs with a changed controller, 7 with a CI above zero (Fig.~\ref{fig:main}A,~B); 5 remain significant after a Holm correction over the 12 pairs. Pooled over the 12 pairs, success rises from 72\% to 80\% (1,718 to 1,931 of 2,400 episodes), a 31\% relative reduction in failures.
On LIBERO-Object, \ctrl{0.5}{0.3} rises from 73\% to 93\% (+20 points [+13.5, +27]); two more classifier seeds give 93.5\% and 91\%.
The exception is LIBERO-10, where \ctrl{0.5}{0.3} drops by 5.5 points [$-13$, +2]. All five second runs (full repeats with new target-body data, a new classifier and a new $w$; three controller bodies and two KUKA arm runs) also gain; two of them were registered after inconclusive first runs.

\textbf{Pre-registered sweep.} Of six more controller bodies registered before data collection (squares in Fig.~\ref{fig:main}A), 3 met our criterion, so our registered family claim (at least 4 of 6) fails (App.~\ref{app:sweep}). Among controller changes, guidance helps most when unguided \pz{} succeeds on a third to two thirds of episodes (Fig.~\ref{fig:gainnaive}).

\subsection{Weak wrists and other faults: mixed}
\label{sec:res-mixed}
Guidance lifts the Kinova arm (10\% wrist) from 70\% to 85\% (+15 points [+7.5, +22.5]) and the KUKA arm (25\% wrist) from 42\% to 52.5\% (3-seed mean), but lowers the Sawyer arm (41\% to 25\%) and the Panda with a 10\% wrist (47\% to 37.5\%); over all 12 original LIBERO-Object bodies (controller changes included), the net gain is only +0.3 points.
The worst case is a locked wrist (68.5\% to 36\%): guided plans ask for wrist roll the joint cannot make, and switching guidance on only after the first grasp restores 67\%.
Where the fault is known, simple rules can win: smoothing on the capped UR5e (85.5\% vs.\ 77.5\%; unguided 84\%), and sending actions planned 4 steps ahead on \ctrl{1}{0.2} (97.5\% at 280 steps).

\subsection{Ablations and guided collection}
\label{sec:res-grad}
\label{sec:res-gsc}
\textbf{The gradient is what helps.}
Reranking and LoRA both ran on the KUKA arm, with reranking also on \ctrl{0.5}{0.3}.
Reranking $N$ unguided samples with the guidance classifier ($N=8$ on the KUKA arm, 4 on \ctrl{0.5}{0.3}) shows no gain (KUKA arm: 35.5\% vs.\ 42\% unguided; \ctrl{0.5}{0.3}: 76\% vs.\ 73\%), while gradient guidance with the same classifiers helps (73.5\% and 93\%; on the KUKA arm, both use the seed-0 guided-collection classifier); \pz{}'s samples at one state differ by only 0.1--0.5 logits.
With the same number of target-body \pz{} queries as the classifier (56 successful episodes), an untuned LoRA fine-tune drops to 29.5\% ($-12.5$ points [$-21$, $-4$]); on a 26\% larger re-collection made with the guided-collection protocol it reaches 41.5\%, against 73.5\% for guidance with the seed-0 guided-collection classifier (App.~\ref{app:baselines}).

\textbf{Guided self-collection.}
On the KUKA arm, the original unguided data had no successes on calibration task 2, but guided \pz{} solved it.
Retraining the classifier with the guided successes from all four calibration tasks raises transfer-task success from 52.5\% to 70.7\% (3-seed means; Fig.~\ref{fig:main}C), against 57.8\% when the same number of extra queries came from unguided \pz{}.
The round was designed to fix a loss on transfer task 3, so this transfer-task set is not fully clean (App.~\ref{app:guided}). A second round added nothing beyond the same number of extra episodes (75.2\% vs.\ 74.5\%), and guided collection did not help \ctrl{0.5}{0.3} on LIBERO-Goal (67\% both).

\section{Related Work}
\label{sec:related}
Our operator is the density-ratio correction of discriminator guidance~\cite{kim2023discriminator} and TGDP~\cite{ouyang2024tgdp}, applied to a frozen model as in classifier guidance~\cite{dhariwal2021diffusion}.
Other steering signals for frozen robot policies encode the task or its success, such as a value function~\cite{nakamoto2024vgps} or a noise-space actor~\cite{wagenmaker2025dsrl}, or a known controller's cost~\cite{gupta2025umionair}; embodiment-transfer methods fine-tune on target demonstrations~\cite{zhang2025ate}, bridge appearance~\cite{chen2024mirage} or need paired rollouts~\cite{jo2026jparc}.
We learn the correction from success-labeled rollouts alone (App.~\ref{app:related}).

\section{Limitations and Conclusion}
\label{sec:limitations}
Embodiment-classifier guidance can steer a frozen \pz{} toward some target bodies, mostly controller changes, but we cannot yet tell which in advance: a pre-registered rule for when to guide failed (App.~\ref{app:stats}).
\textbf{Limitations.}
(i) All results are in simulation, with one policy and one benchmark.
(ii) The method needs success-labeled rollouts on both bodies, and target-body data for each suite.
(iii) The classifier's input carries the end-effector pose, so we have not separated where the bodies go from how they move.
(iv) \pz{} saw the transfer tasks on its source body, and transfer task 3 informed guided collection.
(v) Most results use one classifier and evaluation seed; after a Holm correction, 5 of the 12 controller pairs remain significant; guidance can hurt, and simple rules fix known faults such as a fixed delay better.
(vi) We test one single-body fine-tuned checkpoint; steering a multi-body base policy is untested.
(vii) Guidance adds a classifier gradient at each of the 10 denoising steps, which slows every query; since the classifier is small (0.8M weights versus \pz{}'s 3B), we expect the overhead to be modest, but we did not measure it under controlled conditions.
(viii) Our evidence for data efficiency rests on one fine-tuning baseline: a single untuned LoRA run on one body, trained on task-imbalanced data, whereas $w$ was selected for guidance; a tuned comparison across data budgets and bodies is needed to establish it.

\label{mainbodyend}

\clearpage
\bibliography{refs}

\clearpage
\appendix
\let\origsection\section
\renewcommand{\section}{\FloatBarrier\origsection}
\section{Setup and Protocol}
\label{app:setup}
\textbf{Reporting convention.} Throughout the appendix, success is given as a percentage with the underlying count, e.g.\ 73\% (146/200), and differences are in percentage points (points) with 95\% paired-bootstrap CIs, as in the main text.

\textbf{Simulator and policy.}
All runs use LIBERO~\cite{liu2023libero} on robosuite~\cite{zhu2020robosuite} (MuJoCo physics): one robot arm at a table with household objects.
The policy is \pz{}~\cite{black2024pi0}, a 3-billion-parameter VLA, from Physical Intelligence's public LIBERO checkpoint (JAX), frozen.
Its input is two camera images (third-person and wrist), the robot state and the instruction; its action is 7-D (end-effector translation, rotation and gripper), sent at 20\,Hz.
Each \pz{} query returns a 50-step chunk, of which the robot executes 5 (0.25\,s).
On the source body, LIBERO's operational-space controller (OSC) turns each end-effector command into joint torques.
Each task has 50 fixed starting states (ids 0--49), and success is LIBERO's task predicate on the final scene. Table~\ref{tab:tasks} lists the tasks and Table~\ref{tab:bodies} the bodies.

\begin{table}[ht]
\centering
\scriptsize
\setlength{\tabcolsep}{3pt}
\begin{tabular}{@{}llp{3.95in}@{}}
\toprule
Suite (limit) & Task & Instruction \\
\midrule
\textbf{LIBERO-Object} & 1 (calibration) & pick up the cream cheese and place it in the basket \\
\quad 560 steps & 2 (calibration) & pick up the salad dressing and place it in the basket \\
\quad (native 280) & 4 (calibration) & pick up the ketchup and place it in the basket \\
 & 5 (calibration) & pick up the tomato sauce and place it in the basket \\
 & 0 (transfer) & pick up the alphabet soup and place it in the basket \\
 & 3 (transfer) & pick up the bbq sauce and place it in the basket \\
 & 6 (transfer) & pick up the butter and place it in the basket \\
 & 7 (transfer) & pick up the milk and place it in the basket \\
\midrule
\textbf{LIBERO-Spatial} & 0 (calibration) & pick up the black bowl between the plate and the ramekin and place it on the plate \\
\quad 440 steps & 2 (calibration) & pick up the black bowl from table center and place it on the plate \\
\quad (native 220) & 4 (calibration) & pick up the black bowl in the top drawer of the wooden cabinet and place it on the plate \\
 & 6 (calibration) & pick up the black bowl next to the cookie box and place it on the plate \\
 & 1 (transfer) & pick up the black bowl next to the ramekin and place it on the plate \\
 & 3 (transfer) & pick up the black bowl on the cookie box and place it on the plate \\
 & 5 (transfer) & pick up the black bowl on the ramekin and place it on the plate \\
 & 7 (transfer) & pick up the black bowl on the stove and place it on the plate \\
\midrule
\textbf{LIBERO-Goal} & 0 (calibration) & open the middle drawer of the cabinet \\
\quad 600 steps & 2 (calibration) & put the wine bottle on top of the cabinet \\
\quad (native 300) & 4 (calibration) & put the bowl on top of the cabinet \\
 & 6 (calibration) & put the cream cheese in the bowl \\
 & 1 (transfer) & put the bowl on the stove \\
 & 3 (transfer) & open the top drawer and put the bowl inside \\
 & 5 (transfer) & push the plate to the front of the stove \\
 & 7 (transfer) & turn on the stove \\
\midrule
\textbf{LIBERO-10} & 0 (calibration) & put both the alphabet soup and the tomato sauce in the basket \\
\quad 980 steps & 2 (calibration) & turn on the stove and put the moka pot on it \\
\quad (native 520; & 4 (calibration) & put the white mug on the left plate and put the yellow and white mug on the right plate \\
\quad env.\ cap 1,000) & 6 (calibration) & put the white mug on the plate and put the chocolate pudding to the right of the plate \\
 & 1 (transfer) & put both the cream cheese box and the butter in the basket \\
 & 3 (transfer) & put the black bowl in the bottom drawer of the cabinet and close it \\
 & 5 (transfer) & pick up the book and place it in the back compartment of the caddy \\
 & 7 (transfer) & put both the alphabet soup and the cream cheese box in the basket \\
\bottomrule
\end{tabular}
\caption{\textbf{Tasks and their exact instructions.} Calibration tasks supply target-body data for the classifier and the choice of $w$; all reported results are on transfer tasks, which have no target-body data. \pz{} was trained on all tasks with the source body. The splits were fixed before any data existed, and code guards block transfer task ids from collection and selection.}
\label{tab:tasks}
\end{table}

\begin{table}[ht]
\centering
\scriptsize
\setlength{\tabcolsep}{3pt}
\begin{tabular}{@{}lp{3.75in}@{}}
\toprule
Body & Exact change from the source body \\
\midrule
source body, \ctrl{1}{0} & stock Panda arm, LIBERO's OSC controller at standard stiffness, no delay \\
\midrule
\multicolumn{2}{@{}l}{\emph{Controller changes} (Panda; camera images unchanged). \ctrl{$k$}{$d$}: stiffness scaled by $k$, commands held back by $d$ seconds} \\
\ctrl{1}{0.2} & standard stiffness; commands delayed 4 control steps \\
\ctrl{0.5}{0.2} & stiffness halved; commands delayed 4 steps \\
\ctrl{0.5}{0.3} & stiffness halved; commands delayed 6 steps \\
\ctrl{0.33}{0.3} & stiffness one third; commands delayed 6 steps \\
sweep (App.~\ref{app:sweep}) & \ctrl{0.67}{0.3}, \ctrl{0.5}{0.45}, \ctrl{0.33}{0.15}, \ctrl{0.5}{0.1}, \ctrl{0.75}{0.45}, \ctrl{0.33}{0.45} \\
\midrule
\multicolumn{2}{@{}l}{\emph{Weak wrist} (Panda gripper kept; camera images change for the other arms)} \\
KUKA arm (25\% wrist) & KUKA LBR iiwa 7-joint arm; last three joint motors (wrist) at 25\% strength (gear) \\
Kinova arm (10\% wrist) & Kinova Gen3 arm; wrist motors at 10\% strength \\
Sawyer arm (25\% wrist) & Sawyer arm; wrist motors at 25\% strength \\
Panda (10\% wrist) & the Panda's own wrist motors at 10\% strength \\
sweep (App.~\ref{app:sweep}) & the KUKA arm with wrist motors at 15\% or 50\% strength (both stopped before evaluation) \\
\midrule
\multicolumn{2}{@{}l}{\emph{Other faults}} \\
UR5e (speed cap) & UR5e arm with the Panda gripper, under its own joint-velocity (resolved-rate) controller; joint speed capped at 0.3\,rad/s \\
Panda (slow gripper) & gripper opens and closes 5$\times$ slower \\
worn Panda & weaker shoulder and elbow motors, joint friction, a 10-degree calibration (frame) error and a slow gripper \\
Panda (locked wrist) & wrist-roll joint (joint 7) locked \\
Panda (locked upper arm) & upper-arm joint (joint 3) locked; outside the pilot band, reported separately (App.~\ref{app:failures}) \\
\bottomrule
\end{tabular}
\caption{\textbf{The target bodies.} Stiffness is the gain of the end-effector controller; a delay holds each command back for a fixed number of 20\,Hz control steps (2, 3, 4, 6 and 9 steps are 0.1, 0.15, 0.2, 0.3 and 0.45\,s).}
\label{tab:bodies}
\end{table}

\textbf{Is the KUKA arm's loss the arm or the wrist?}
A KUKA iiwa with a full-strength wrist succeeds on 87.5\% (70/80) of transfer-task pilot episodes, against 36.3\% (29/80) with the 25\% wrist on the same starting states, so the arm swap alone does not explain the loss. That the weak wrist causes most of it is an inference: restoring the wrist within the same run rescued about 40\% of the failures.

\textbf{Classifier data.}
The embodiment classifier's target-body data come from unguided \pz{} on the calibration tasks, 2,500 \pz{} queries per task, keeping successful episodes; source-body data come from successful source-body rollouts on the same tasks. Each suite uses only its own calibration tasks.
On LIBERO-Object, a 560-step episode uses at most 112 queries (5 executed actions each), so the budget gives at least 22 episodes per task, and more when episodes end early; this is the ``about 23 rollouts per calibration task'' of the main text.
Chunks are built as \pz{}'s input transform builds them: the next 50 commanded actions, with the current state subtracted from dimensions 0--5 (end-effector position and axis-angle rotation), then z-scored with the checkpoint's statistics.
The subtraction means that every chunk carries the current end-effector pose as a constant offset.
A classifier for \ctrl{1}{0.2} trained on one task reached validation AUC 0.998 on that task but 0.65 on transfer tasks; trained on four tasks, its transfer-task AUC was 0.919, 0.907 and 0.913 over three seeds with fresh data.

\textbf{Choosing $w$.}
We try $w\in\{1,2,4\}$ on the calibration tasks (starting states 0--19, 80 episodes, evaluation seed 0) and keep the one with the most successes, breaking ties toward the smaller $w$. For \ctrl{0.5}{0.3}, \ctrl{1}{0.2} and the UR5e, $w$ was chosen at 280 steps and evaluated at 560 steps, as stated in their registrations.
The choice is mostly within noise: in 31 of 38 runs with saved selection scores, the chosen $w$ won by at most 5 points (4/80), about one standard deviation of such a difference.

\textbf{Episode limits.}
If the target body does the same motions $s$ times slower, success within a fixed limit $H$ drops by $F_A(H)-F_A(H/s)$, where $F_A(h)$ is the source body's chance of success within $h$ steps; slowness alone would then count as failure. We therefore use about twice LIBERO's native limits (Table~\ref{tab:tasks}). On \ctrl{1}{0.2}, the gain read at 280 steps was 22.5 points, against 11.75 points on the same episodes at 560 steps (App.~\ref{app:robust}).

\subsection{Statistics and pre-registration}
\label{app:stats}
\begin{itemize}
\item \textbf{Pairing.} Every condition uses the same 50 starting states per task and the same environment seed. Episodes are not bit-reproducible across runs, so ``paired'' means the same starting state, not the same random stream.
\item \textbf{Intervals and tests.} Per condition: success and a 95\% Wilson interval. Differences: paired bootstrap over (task, starting state) pairs, 10,000 resamples, percentile interval, and an exact two-sided McNemar test on discordant pairs. The bootstrap was run separately for the text and for the per-task tables, so a CI endpoint can differ by 0.5 points between them. All numbers were recomputed from per-episode records by a separate script.
\item \textbf{Multiple comparisons.} Over the 12 controller-change suite--body pairs, a Holm correction at $\alpha=0.05$ leaves 5 significant: Object \ctrl{0.5}{0.3} and \ctrl{0.33}{0.3}, Spatial \ctrl{0.5}{0.3} and \ctrl{0.33}{0.3}, and Goal \ctrl{0.33}{0.3}. It uses exact McNemar $p$-values where reported and otherwise $p$-values approximated from the paired-bootstrap CIs (normal approximation). Other rows are uncorrected; App.~\ref{app:reuse} applies a Bonferroni correction over the four reuse tests.
\item \textbf{Registration.} Each run's conditions, horizon, seeds and criteria were committed to version control before its data existed. The usual criterion is a gain of at least 5 points (some runs: 10 points) with a paired CI above zero; a gain below 1.5 points counts as no effect, and anything in between, or a CI touching zero, is inconclusive.
\item \textbf{Body screen.} A body entered a full run only if unguided \pz{} succeeded on 30--75\% of an 80-episode transfer-task pilot. \ctrl{0.5}{0.2} entered after the band was widened from 30--70\% to 30--75\%, after its pilot was seen. The sweep bodies had no pilot; registered no-room and floor rules took its place.
\item \textbf{Guide-or-not rule (failed).} We registered a rule to guide only if the best guided calibration-task success is at least 10 points (8/80) above unguided. On 10 bodies with known outcomes it had to be right on at least 7 of the 8 clear-cut ones, beat always-guiding by at least 67 successes in total, and block at most one body that gains. It was right on 5 of 8: it blocked all three harmful bodies (Panda (10\% wrist), Sawyer arm, locked wrist) but also three that gain (Kinova arm, KUKA arm, \ctrl{1}{0.2}), and beat always-guiding by 50.5 successes. A 5-point margin would have been right on 7 of 8, but it was chosen after seeing the results, so we do not use it. Calibration-task success on 80 episodes is too noisy to separate a real gain from none, so the method always guides.
\item \textbf{Deviations.} Two stopping rules for the sweep were added after its first five bodies had run, and they stopped two bodies before evaluation (App.~\ref{app:sweep}). The second LIBERO-Goal runs of the KUKA arm and \ctrl{0.5}{0.3} were registered after their first runs came out inconclusive. Analyses labeled descriptive had no registered criterion.
\end{itemize}

\section{Method Details and Theory}
\label{app:theory}
\label{app:method}

\begin{figure}[ht]
\centering
\includegraphics[width=0.85\linewidth]{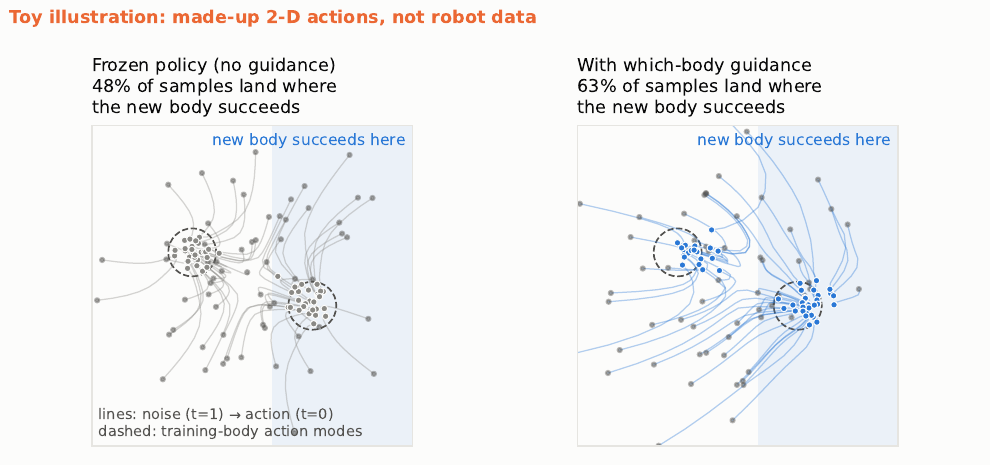}
\caption{\textbf{Toy illustration with made-up 2-D data (not robot data).} Each point is a 2-D ``action''. The source body's actions form two clusters (dashed circles); the target body succeeds only in the shaded region. Lines show 60 samples moving from noise ($t=1$) to actions ($t=0$) under the exact flow of the source-body data. Left: the frozen flow puts 48\% of samples in the shaded region. Right: adding a capped classifier gradient (here a fixed linear logit that prefers larger $x_1$) puts 63\% there. Guidance mostly moves samples toward the source-body cluster that also works on the target body.}
\label{fig:toy}
\end{figure}

\textbf{Intuition.}
The embodiment classifier learns which chunk shapes separate the target body's successful chunks from the source body's chunks, and its gradient points toward chunks that look more like the former.
Each nudge therefore bends the plan toward motions that worked on the target body, while \pz{}'s velocity field still provides the task-level plan (which object to reach and where to place it) (we did not measure which features the classifier uses).
This picture is consistent with the failures in App.~\ref{app:failures}: if no plan of a useful shape exists (a locked joint), the nudge asks for motion the robot cannot make; if the fault is timing (a slow gripper), plan shape is not the cause; and where the classifier saw no target-body data, its gradient is a guess.
Figure~\ref{fig:toy} shows a toy 2-D example.

\textbf{When guided self-collection runs.}
The round runs only if guided \pz{} solves a calibration task on which $D_B$ has few or no successes.
The KUKA arm (25\% wrist) met this condition; the Sawyer arm did not (0 guided task-2 successes); \ctrl{0.5}{0.3} on LIBERO-Goal ran the round, which did not help (App.~\ref{app:guided}).

Steps marked \textbf{[heuristic]} below are not exact.

\subsection{Flow path, velocity and score}
\pz{} uses $x_t=t\varepsilon+(1-t)a$ and target $u=\varepsilon-a$, trained with $t\sim0.999\,\mathrm{Beta}(1.5,1)+0.001$.
The best network is the conditional mean $v(x,t,o)=\E[\varepsilon-a\mid x_t=x,o]$.
Let $\hat a=\E[a\mid x_t=x]$ and $\hat\varepsilon=\E[\varepsilon\mid x_t=x]$, so $x=t\hat\varepsilon+(1-t)\hat a$.
\begin{enumerate}
\item Velocity: $v=\hat\varepsilon-\hat a=(\hat\varepsilon-x)/(1-t)$.
\item Score (Tweedie's formula): $p_t(x)=\int p(a)\,\mathcal N(x;(1-t)a,t^2I)\,da$ gives $s_t(x)=\nabla_x\log p_t(x)=-\hat\varepsilon/t$.
\item Combining: $v(x,t)=-\big(x+t\,s_t(x)\big)/(1-t)$.
\end{enumerate}
The $x$ term is the same for every data distribution, so for the two chunk distributions $q^A$ and $q^B$ on the same path
\begin{equation}
v^B(x,t)-v^A(x,t)=-\frac{t}{1-t}\,\nabla_x\log\frac{q^B_t(x)}{q^A_t(x)} .
\end{equation}
This identity is exact for all $t\in(0,1)$, and with the exact ratio the guided ODE carries noise to $q^B$.
A 1-D check ($p_A=\mathcal N(0,1)$, $p_B=\mathcal N(1.5,0.5^2)$, 1,000 Euler steps) gave endpoint mean 1.5001 and s.d.\ 0.4999.
\emph{Sign check.} The sampler moves $x$ by $\Delta\,\delta v$ with $\Delta<0$, so the move is $|\Delta|\,\frac{t}{1-t}\nabla\ell$: uphill in $\ell$, toward the target body.

\subsection{The intended distribution and what an action-only classifier estimates}
$D_X$ holds chunks from successful episodes of unguided \pz{} on body $X$, on the calibration tasks.
Let $\mathcal S$ be success and $\rho_X(o)$ the observations visited on body $X$.
Treating episode success as a property of each chunk \textbf{[heuristic]}, the chunk density in $D_X$ is $\pi_X(a\mid o)=p_A(a\mid o)P_X(\mathcal S\mid o,a)/P_X(\mathcal S\mid o)$.
The intended distribution is $\pi_B(a\mid o)=p_A(a\mid o)\,r(a,o)/Z(o)$ with density ratio $r=\pi_B/p_A$, that is, \pz{} reweighted by the probability of success on the target body.
The classifier never sees $o$, so it can only compare $q_X(a)=\int\pi_X(a\mid o)\rho_X(o)\,do$.
Its optimum equals $\log r(a,o)$, up to a term in $o$ only, under these assumptions:
\begin{enumerate}
\item[A0] $\pi_A\approx p_A$: \pz{} nearly always succeeds on the source body.
\item[A1] $\rho_B=\rho_A$: both bodies visit the same observations. False in general; a slow body sees different states.
\item[A2] The noised ratio $R_t(x,o)=p^B_t(x\mid o)/p^A_t(x\mid o)$ is free of $o$ at every $t$.
\item[A3] Each chunk of a successful episode counts as a success chunk.
\item[A4] Planned chunks (what guidance acts on) and executed windows (what the classifier trains on) have the same distribution.
\item[A5] The ratio learned on calibration tasks also holds on transfer tasks.
\end{enumerate}
Without A1 and A2, $q^B_t(x)/q^A_t(x)=\E_{o\sim\rho_A(o\mid x_t=x)}\big[\tfrac{\rho_B(o)}{\rho_A(o)}R_t(x,o)\big]$, so every state-dependent part of the correction is lost and only the average change in chunk shape remains.
One exception: chunks in \pz{}'s action space have the current end-effector pose subtracted (App.~\ref{app:setup}), so the pose part of $o$ reaches the classifier, and a violation of A1 in where the bodies go can be read off the chunks.

\subsection{The noise-conditioned classifier}
The embodiment classifier's training loss, with the target body as the positive class and 128 chunks per class in each batch, is
\begin{equation}
L(\phi)=\E_{a\sim D_A,t,\varepsilon}\,\mathrm{sp}\big(\ell_\phi(x_t,t)\big)+\E_{a\sim D_B,t,\varepsilon}\,\mathrm{sp}\big(-\ell_\phi(x_t,t)\big),\quad \mathrm{sp}(z)=\log(1+e^z),
\label{eq:loss}
\end{equation}
with $t\sim U(0,1)$.
For fixed $(x,t)$, the pointwise loss $q^A_t\,\mathrm{sp}(\ell)+q^B_t\,\mathrm{sp}(-\ell)$ is minimized at $\ell^\star(x,t)=\log q^B_t(x)/q^A_t(x)$; the distribution of $t$ only weights noise levels.
Noise conditioning is needed because the noised ratio is an average of the clean ratio $r_0$, $r_t(x)=\E_{a\sim q_A(a\mid x_t=x)}[r_0(a)]$, and the log of an average is not the average of the log.
Guidance acts on the 7 real action dimensions of \pz{}'s 32 padded dimensions, at all 10 flow steps.
The network has 4 transformer blocks of width 128 (about 0.8M weights) and mixes information across the 50 chunk steps.

\subsection{Gaps between the implementation and the exact correction}
\begin{itemize}
\item \textbf{The clamp at $t=1$.} The implementation uses the factor $1/0.05=20$ at the first step. A perfect classifier has zero gradient there, because both noised marginals are $\mathcal N(0,I)$, while the exact correction at $t=1$ is a mean shift $-(\mu_B-\mu_A)$. We did not measure the size of this gap.
\item \textbf{The weight $w\neq1$} \textbf{[heuristic]}. The field then targets intermediate densities $\propto p^A_t r_t^{\,w}$, which are not the noised versions of $p_A r^w$; this is the known gap of guidance with a scale.
\item \textbf{The cap} \textbf{[heuristic]}. It breaks exactness and is often active (71\% of guidance calls on task 0 for the KUKA arm).
\item \textbf{Discretization and network error.} Ten Euler steps and the network's own error are not bounded here.
\end{itemize}
With $w>1$ and the cap, we therefore sample from no known distribution; we call the result ``guided'', not ``samples of $\pi_B$''.

\section{Full Results}
\label{app:results}
\begin{figure}[ht]
\centering
\includegraphics[width=0.85\linewidth]{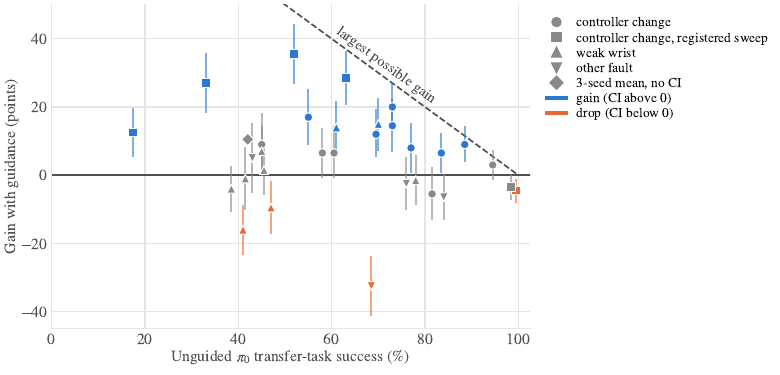}
\caption{\textbf{Gain vs.\ unguided success} for every first run in Table~\ref{tab:main} and the six evaluated sweep bodies (Table~\ref{tab:sweep}), with 95\% paired-bootstrap CIs. The dashed line is the largest possible gain. Gains appear across a wide range of starting success; near the ceiling there is no room, and the largest drops sit at low or middle success.}
\label{fig:gainnaive}
\end{figure}

Table~\ref{tab:main} lists every body-by-suite result, Fig.~\ref{fig:gainnaive} plots each gain against unguided success, and Tables~\ref{tab:body_task} and~\ref{tab:body_task_suites} break the results down by task.

\begin{table}[ht]
\centering
\scriptsize
\setlength{\tabcolsep}{2.4pt}
\begin{tabular}[t]{@{}lrrl@{}}
\toprule
\multicolumn{4}{@{}l}{\textit{LIBERO-Object, 560 steps, transfer tasks 0, 3, 6, 7}}\\
Body & Ung. \% (n) & Guid. \% (n) & Diff, points [95\% CI] \\
\midrule
\ctrl{0.5}{0.3} & 73 (146) & 93 (186) & +20 [+13.5, +27] \\
\quad classifier seed 1 & 73 (146) & 93.5 (187) & +20.5 [+13.5, +27.5] \\
\quad classifier seed 2 ($w=4$) & 73 (146) & 91 (182) & +18 [+11, +25] \\
\quad eval seed 1 & 72 (144) & 96 (192) & +24 [+17.5, +30.5] \\
\ctrl{0.33}{0.3} & 88.5 (177) & 97.5 (195) & +9 [+4, +14] \\
\ctrl{1}{0.2} & 77 (154) & 85 (170) & +8 [+1, +15] \\
\quad eval seed 1 & 72.5 (145) & 88 (176) & +15.5 [+8, +23] \\
\ctrl{0.5}{0.2} & 94.5 (189) & 97.5 (195) & +3 [$-$1, +7] \\
KUKA arm (25\% wrist), 3 seeds & 42 (84) & 52.5 (105) & +10.5 \\
\quad + guided collection & 42 (84) & 70.7 (141.3) & +28.7 \\
Kinova arm (10\% wrist) & 70 (140) & 85 (170) & +15 [+7.5, +22.5] \\
worn Panda & 43 (86) & 48 (96) & +5 [$-$5, +15] \\
UR5e (speed cap) & 84 (168) & 77.5 (155) & $-$6.5 [$-$13, 0] \\
Panda (slow gripper) & 76 (152) & 73.5 (147) & $-$2.5 [$-$10, +5] \\
Panda (10\% wrist) & 47 (94) & 37.5 (75) & $-$9.5 [$-$17, $-$2] \\
Sawyer arm (25\% wrist) & 41 (82) & 25 (50) & $-$16 [$-$23, $-$9] \\
Panda (locked wrist) & 68.5 (137) & 36 (72) & $-$32.5 [$-$41, $-$24] \\
\bottomrule
\end{tabular}\hfill
\begin{tabular}[t]{@{}lrrll@{}}
\toprule
\multicolumn{5}{@{}l}{\textit{Other suites, 2$\times$ native limit, transfer tasks 1, 3, 5, 7}}\\
Suite: body & Ung. \% (n) & Guid. \% (n) & Diff, points [95\% CI] & 2nd; pool \\
\midrule
Spatial: \ctrl{0.5}{0.3} & 55 (110) & 72 (144) & +17 [+9, +25] & +14.5; +18.5 \\
Spatial: \ctrl{0.33}{0.3} & 73 (146) & 87.5 (175) & +14.5 [+7, +22] & --; +14 \\
Spatial: KUKA (25\%) & 61 (122) & 75 (150) & +14 [+6.5, +21.5] & +12; +13.75 \\
Spatial: \ctrl{1}{0.2} & 45 (90) & 54 (108) & +9 [0, +18] & --; +5.25 \\
Spatial: \ctrl{0.5}{0.2} & 83.5 (167) & 90 (180) & +6.5 [+1, +12] &  \\
Spatial: Kinova (10\%) & 78 (156) & 76.5 (153) & $-$1.5 [$-$8.5, +5.5] &  \\
Goal: \ctrl{0.33}{0.3} & 69.5 (139) & 81.5 (163) & +12 [+5.5, +19] & +7; +8 \\
Goal: KUKA (25\%) & 45 (90) & 52 (104) & +7 [0, +14] & +13.5$^\dagger$; +8.75 \\
Goal: \ctrl{0.5}{0.3} & 60.5 (121) & 67 (134) & +6.5 [$-$0.5, +14] & +12.5$^\dagger$; +5 \\
Goal: \ctrl{1}{0.2} & 58 (116) & 64.5 (129) & +6.5 [$-$0.5, +13.5] &  \\
Goal: Kinova (10\%) & 45.5 (91) & 47 (94) & +1.5 [$-$5.5, +9] &  \\
L-10: \ctrl{0.5}{0.3} & 81.5 (163) & 76 (152) & $-$5.5 [$-$13, +2] &  \\
L-10: Kinova (10\%) & 38.5 (77) & 34.5 (69) & $-$4 [$-$10.5, +2.5] &  \\
L-10: KUKA (25\%) & 41.5 (83) & 40.5 (81) & $-$1 [$-$10, +8] &  \\
\bottomrule
\end{tabular}
\caption{\textbf{Transfer-task success, unguided (Ung.) vs.\ guided (Guid.) \pz{}}, in \% with successes out of 200 in parentheses. One evaluation seed and classifier per row unless stated. KUKA arm: classifier seeds give 57\% (114/200), 52\% (104/200) and 48.5\% (97/200); the guided-collection row is the 3-seed mean (App.~\ref{app:guided}). 2nd: the gain of a second run with fresh target-body data, a new classifier and a new $w$; pool: the first run's gain averaged over two evaluation seeds (all in points). $^\dagger$Registered after the first run was inconclusive; these second runs go from 43.5\% (87/200) to 57\% (114/200) and from 61\% (122/200) to 73.5\% (147/200). L-10: LIBERO-10. The sweep bodies are in Table~\ref{tab:sweep}.}
\label{tab:main}
\end{table}

\begingroup
\let\origtabular\tabular
\let\endorigtabular\endtabular
\renewenvironment{tabular}{\begin{adjustbox}{max width=\textwidth}\origtabular}{\endorigtabular\end{adjustbox}}
\begin{table*}[t]
\centering
\small
\begin{tabular}{lccccclc}
\toprule
Body & Task 0 & Task 3 & Task 6 & Task 7 & \textcolor{gray}{Task 1} & Total \% (n/200) & Diff, points [95\% CI] \\
\midrule
\ctrl{0.5}{0.3} & \textbf{78 / 94} & \textbf{52 / 84} & \textbf{88 / 100} & \textbf{74 / 94} & \textcolor{gray}{82 / 90} &  73 / 93 (146 / 186) & +20 [+13.5, +27] \\
\ctrl{0.5}{0.3} (eval seed 1) & \textbf{76 / 100} & \textbf{48 / 94} & 96 / 100 & \textbf{68 / 90} & \textcolor{gray}{76 / 84} &  72 / 96 (144 / 192) & +24 [+17.5, +30.5] \\
\ctrl{0.33}{0.3} & 94 / 100 & \textbf{72 / 92} & 92 / 100 & 96 / 98 & \textcolor{gray}{86 / 100} &  88.5 / 97.5 (177 / 195) & +9 [+4, +14] \\
\ctrl{1}{0.2} & 82 / 90 & \textbf{56 / 76} & 98 / 100 & 72 / 74 & \textcolor{gray}{76 / 82} &  77 / 85 (154 / 170) & +8 [+1, +15] \\
\ctrl{1}{0.2} (eval seed 1) & \textbf{80 / 92} & \textbf{40 / 76} & 94 / 100 & 76 / 84 & \textcolor{gray}{72 / 72} &  72.5 / 88 (145 / 176) & +15.5 [+8, +23] \\
UR5e (speed cap) & 80 / 80 & 94 / 78 & 72 / 66 & 90 / 86 & \textcolor{gray}{100 / 92} &  84 / 77.5 (168 / 155) & $-$6.5 [$-$13, 0] \\
\ctrl{0.5}{0.2} & 92 / 94 & 92 / 100 & 100 / 100 & 94 / 96 & \textcolor{gray}{92 / 94} &  94.5 / 97.5 (189 / 195) & +3 [$-$1, +7] \\
KUKA arm (25\% wrist) & \textbf{40 / 58} & 54 / 42 & \textbf{26 / 56} & \textbf{48 / 72} & \textcolor{gray}{68 / 62} &  42 / 57 (84 / 114) & +15 [+5, +25] \\
KUKA arm (25\% wrist) (eval seed 1) & \textbf{40 / 62} & 58 / 48 & \textbf{28 / 56} & \textbf{52 / 74} & \textcolor{gray}{not run} &  44.5 / 60 (89 / 120) & +15.5 [+6, +24.5] \\
Kinova arm (10\% wrist) & \textbf{64 / 80} & \textbf{62 / 86} & \textbf{58 / 84} & 96 / 90 & \textcolor{gray}{80 / 76} &  70 / 85 (140 / 170) & +15 [+7.5, +22.5] \\
Panda (10\% wrist) & 72 / 60 & 76 / 64 & 0 / 2 & 40 / 24 & \textcolor{gray}{60 / 42} &  47 / 37.5 (94 / 75) & $-$9.5 [$-$17, $-$2] \\
Sawyer arm (25\% wrist) & 62 / 52 & 16 / 4 & 46 / 16 & 40 / 28 & \textcolor{gray}{60 / 72} &  41 / 25 (82 / 50) & $-$16 [$-$23, $-$9] \\
Panda (slow gripper) & 80 / 64 & 84 / 68 & \textbf{42 / 68} & 98 / 94 & \textcolor{gray}{60 / 34} &  76 / 73.5 (152 / 147) & $-$2.5 [$-$10, +5] \\
worn Panda & \textbf{44 / 62} & 84 / 46 & \textbf{22 / 54} & 22 / 30 & \textcolor{gray}{72 / 80} &  43 / 48 (86 / 96) & +5 [$-$5, +15] \\
Panda (locked wrist) & 50 / 20 & 86 / 42 & 82 / 40 & 56 / 42 & \textcolor{gray}{100 / 94} &  68.5 / 36 (137 / 72) & $-$32.5 [$-$41, $-$24] \\
Panda (locked upper arm)$^\ddagger$ & 0 / 0 & \textbf{44 / 84} & 0 / 0 & 0 / 0 & \textcolor{gray}{34 / 44} &  11 / 21 (22 / 42) & +10 [+5.5, +15] \\
\bottomrule
\end{tabular}
\caption{\textbf{Per-task results, LIBERO-Object} (560 steps, starting states 0--49). Cells: unguided / guided success in \% of 50 episodes; bold: guided at least 10 points above unguided. Grey: calibration task 1 (not a transfer task). Totals are over transfer tasks 0, 3, 6 and 7 (200 episodes), in \% with counts in parentheses, and the paired difference in points with its 95\% paired-bootstrap CI. $^\ddagger$Outside the pilot band (App.~\ref{app:failures}).}
\label{tab:body_task}
\end{table*}

\begin{table*}[t]
\centering
\small
\begin{tabular}{llccccclc}
\toprule
Suite & Body & Task 1 & Task 3 & Task 5 & Task 7 & \textcolor{gray}{Task 0} & Total \% (n/200) & Diff, points [95\% CI] \\
\midrule
LIBERO-Goal & \ctrl{0.5}{0.3} & 84 / 92 & \textbf{18 / 38} & 42 / 40 & 98 / 98 & \textcolor{gray}{96 / 82} &  60.5 / 67 (121 / 134) & +6.5 [$-$0.5, +14] \\
LIBERO-Goal & Kinova arm (10\% wrist) & 66 / 66 & 22 / 24 & 6 / 10 & 88 / 88 & \textcolor{gray}{0 / 0} &  45.5 / 47 (91 / 94) & +1.5 [$-$6, +8.5] \\
LIBERO-Spatial & \ctrl{0.5}{0.3} & \textbf{60 / 92} & \textbf{20 / 58} & 46 / 40 & 94 / 98 & \textcolor{gray}{54 / 38} &  55 / 72 (110 / 144) & +17 [+9, +25.5] \\
LIBERO-10 & \ctrl{0.5}{0.3} & 76 / 80 & 92 / 86 & 86 / 84 & 72 / 54 & \textcolor{gray}{48 / 70} &  81.5 / 76 (163 / 152) & $-$5.5 [$-$13, +2] \\
LIBERO-Goal & KUKA arm (25\% wrist) & 76 / 66 & 28 / 30 & \textbf{2 / 12} & \textbf{74 / 100} & \textcolor{gray}{0 / 6} &  45 / 52 (90 / 104) & +7 [0, +14] \\
LIBERO-Goal & \ctrl{1}{0.2} & \textbf{76 / 88} & \textbf{24 / 38} & 32 / 34 & 100 / 98 & \textcolor{gray}{96 / 96} &  58 / 64.5 (116 / 129) & +6.5 [$-$0.5, +13.5] \\
LIBERO-Goal & \ctrl{0.33}{0.3} & 96 / 94 & \textbf{40 / 58} & \textbf{42 / 74} & 100 / 100 & \textcolor{gray}{94 / 90} &  69.5 / 81.5 (139 / 163) & +12 [+5.5, +19] \\
LIBERO-Spatial & Kinova arm (10\% wrist) & 62 / 68 & 90 / 98 & 78 / 62 & 82 / 78 & \textcolor{gray}{88 / 74} &  78 / 76.5 (156 / 153) & $-$1.5 [$-$8.5, +5.5] \\
LIBERO-Spatial & KUKA arm (25\% wrist) & \textbf{46 / 80} & \textbf{84 / 94} & 60 / 54 & \textbf{54 / 72} & \textcolor{gray}{74 / 76} &  61 / 75 (122 / 150) & +14 [+6.5, +21.5] \\
LIBERO-Spatial & \ctrl{1}{0.2} & 56 / 50 & \textbf{14 / 46} & 26 / 32 & 84 / 88 & \textcolor{gray}{60 / 48} &  45 / 54 (90 / 108) & +9 [0, +18] \\
LIBERO-10 & Kinova arm (10\% wrist) & 68 / 70 & 4 / 12 & 16 / 12 & 66 / 44 & \textcolor{gray}{42 / 40} &  38.5 / 34.5 (77 / 69) & $-$4 [$-$10.5, +2.5] \\
LIBERO-10 & KUKA arm (25\% wrist) & 50 / 40 & 62 / 54 & \textbf{8 / 44} & 46 / 24 & \textcolor{gray}{26 / 16} &  41.5 / 40.5 (83 / 81) & $-$1 [$-$10, +8] \\
LIBERO-Spatial & \ctrl{0.33}{0.3} & \textbf{56 / 86} & \textbf{56 / 90} & 84 / 82 & 96 / 92 & \textcolor{gray}{88 / 52} &  73 / 87.5 (146 / 175) & +14.5 [+7, +22] \\
LIBERO-Spatial & \ctrl{0.5}{0.2} & \textbf{84 / 96} & \textbf{64 / 80} & 88 / 86 & 98 / 98 & \textcolor{gray}{86 / 98} &  83.5 / 90 (167 / 180) & +6.5 [+1, +12] \\
LIBERO-Spatial & worn Panda$^\dagger$ & 100 / 78 & 94 / 86 & 92 / 86 & 86 / 74 & \textcolor{gray}{98 / 98} &  93 / 81 (186 / 162) & $-$12 [$-$18.5, $-$5.5] \\
LIBERO-Spatial & \ctrl{0.5}{0.3} (second run) & \textbf{64 / 84} & \textbf{20 / 46} & 40 / 46 & 88 / 94 & \textcolor{gray}{64 / 44} &  53 / 67.5 (106 / 135) & +14.5 [+6.5, +22.5] \\
LIBERO-Spatial & KUKA arm (25\% wrist) (second run) & 64 / 72 & 82 / 88 & 60 / 52 & \textbf{48 / 90} & \textcolor{gray}{74 / 66} &  63.5 / 75.5 (127 / 151) & +12 [+4, +20] \\
LIBERO-Goal & \ctrl{0.33}{0.3} (second run) & 96 / 98 & 50 / 52 & \textbf{56 / 80} & 98 / 98 & \textcolor{gray}{94 / 82} &  75 / 82 (150 / 164) & +7 [+1, +13.5] \\
LIBERO-Spatial & \ctrl{0.5}{0.3} (eval seed 1) & \textbf{58 / 86} & \textbf{20 / 66} & 42 / 42 & 90 / 96 & \textcolor{gray}{not run} &  52.5 / 72.5 (105 / 145) & +20 [+11.5, +28.5] \\
LIBERO-Spatial & KUKA arm (25\% wrist) (eval seed 1) & \textbf{56 / 68} & \textbf{76 / 96} & 56 / 44 & \textbf{54 / 88} & \textcolor{gray}{not run} &  60.5 / 74 (121 / 148) & +13.5 [+5.5, +21.5] \\
LIBERO-Spatial & \ctrl{0.33}{0.3} (eval seed 1) & \textbf{60 / 98} & \textbf{58 / 96} & 96 / 80 & 94 / 88 & \textcolor{gray}{not run} &  77 / 90.5 (154 / 181) & +13.5 [+6.5, +20.5] \\
LIBERO-Spatial & \ctrl{1}{0.2} (eval seed 1) & 64 / 46 & \textbf{16 / 44} & 32 / 18 & \textbf{86 / 96} & \textcolor{gray}{not run} &  49.5 / 51 (99 / 102) & +1.5 [$-$7.5, +10.5] \\
LIBERO-Goal & KUKA arm (25\% wrist) (eval seed 1) & 78 / 72 & 32 / 38 & \textbf{2 / 16} & \textbf{68 / 96} & \textcolor{gray}{not run} &  45 / 55.5 (90 / 111) & +10.5 [+3.5, +17.5] \\
LIBERO-Goal & \ctrl{0.33}{0.3} (eval seed 1) & 96 / 90 & \textbf{38 / 52} & \textbf{64 / 74} & 100 / 98 & \textcolor{gray}{not run} &  74.5 / 78.5 (149 / 157) & +4 [$-$2.5, +10.5] \\
LIBERO-Goal & \ctrl{0.5}{0.3} (eval seed 1) & 88 / 76 & \textbf{16 / 46} & 42 / 40 & 100 / 98 & \textcolor{gray}{not run} &  61.5 / 65 (123 / 130) & +3.5 [$-$3.5, +11] \\
\bottomrule
\end{tabular}
\caption{\textbf{Per-task results, other suites} (2$\times$ the native horizon; transfer tasks 1, 3, 5 and 7). Cells as in Table~\ref{tab:body_task}; grey: calibration task 0. $^\dagger$One run, recomputed after the other rows; unguided \pz{} is already at 93\%, so by the registered rule this body has no room to gain.}
\label{tab:body_task_suites}
\end{table*}

\endgroup

\subsection{Registered sweep of eight more LIBERO-Object bodies}
\label{app:sweep}
After the first twelve LIBERO-Object bodies, we registered eight target bodies (Table~\ref{tab:sweep}) on 2026-10-13, before any of their data existed, fixing the list, its order, the recipe and the criteria, and committing to report every body.
The recipe is the one used for the other LIBERO-Object bodies (calibration tasks 1, 2, 4 and 5; classifier seed 0; $w\in\{1,2,4\}$; transfer tasks 0, 3, 6 and 7; evaluation seed 0; 560 steps).
Criteria: a gain of at least 5 points with a paired CI above zero; below 1.5 points is no effect; unguided success of at least 90\% means no room, and below 30\% means floor (reported only).
Family claim: if at least 4 of the 6 controller bodies meet the criterion, guidance helps across this family.

\begin{table}[!ht]
\centering
\scriptsize
\setlength{\tabcolsep}{2.6pt}
\begin{adjustbox}{max width=\textwidth}
\begin{tabular}{@{}rllll>{\raggedright\arraybackslash}p{1.75cm}@{}}
\toprule
\# & Body & Unguided [Wilson] & Guided [Wilson] ($w$) & Diff, points [95\% CI]; McNemar $p$ & Reading \\
\midrule
1 & \ctrl{0.67}{0.3} & 63\% (126/200) [56, 69] & 91.5\% (183/200) [87, 95] (2) & +28.5 [+21, +36]; $3\times10^{-12}$ & meets criterion \\
2 & \ctrl{0.5}{0.45} & 33\% (66/200) [27, 40] & 60\% (120/200) [53, 67] (1) & +27 [+18.5, +35.5]; $1\times10^{-8}$ & meets criterion \\
3 & \ctrl{0.33}{0.15} & 99.5\% (199/200) [97, 100] & 95\% (190/200) [91, 97] (1) & $-4.5$ [$-8$, $-1.5$]; 0.012 & no room; significant drop \\
4 & \ctrl{0.5}{0.1} & 98.5\% (197/200) [96, 99] & 95\% (190/200) [91, 97] (1) & $-3.5$ [$-7$, 0]; 0.09 & no room \\
5 & \ctrl{0.75}{0.45} & 17.5\% (35/200) [13, 23] & 30\% (60/200) [24, 37] (1) & +12.5 [+5.5, +19.5]; 0.0013 & floor, reported only \\
6 & KUKA arm, 15\% wrist & -- & -- & -- & stopped$^a$ \\
7 & KUKA arm, 50\% wrist & -- & -- & -- & stopped$^b$ \\
8 & \ctrl{0.33}{0.45} & 52\% (104/200) [45, 59] & 87.5\% (175/200) [82, 91] (2) & +35.5 [+27, +44]; $7\times10^{-14}$ & meets criterion \\
\bottomrule
\end{tabular}
\end{adjustbox}
\caption{\textbf{Registered sweep of eight new LIBERO-Object bodies}, in registered order; Wilson intervals in \%. 200 paired transfer-task episodes per body; one data set, classifier seed and evaluation seed each. Tally: 3 meet the criterion, 2 have no room (1 significant drop), 1 is below the floor and 2 were stopped before evaluation. $^a$Stopped by a rule added after bodies 1--5: unguided success during collection was 11.1\% (11/99), below 25\%. $^b$Stopped by a second rule added after bodies 1--5: guided calibration-task success, 76.3\% (61/80), was 18.2 points below unguided success during collection (94.5\%). Both stopped bodies count as not meeting the criterion.}
\label{tab:sweep}
\end{table}

\textbf{Result.}
Three of the eight bodies meet the criterion, and the family claim fails (3 of 6 controller bodies); the two stopped bodies are KUKA arms, so they do not enter the family count.
The three gains (27 to 35.5 points) are the largest of any body without guided collection, and with $p\leq10^{-8}$ they would survive a Bonferroni correction over all rows in this paper; the significant drop on \ctrl{0.33}{0.15} ($p=0.012$; 10 pairs lost success, 1 gained) would not.
Task 3 carries much of several gains: on \ctrl{0.67}{0.3} it rises from 34\% (17/50) to 96\% (48/50), and on \ctrl{0.5}{0.45} from 4\% (2/50) to 60\% (30/50); on \ctrl{0.5}{0.45} and \ctrl{0.33}{0.45}, the gains without task 3 are still +17.3 [+7.3, +27.3] and +30 [+20.7, +39.3] points. The below-floor \ctrl{0.75}{0.45} gains mostly on task 6 (+5.3 [$-2$, +12.7] points without it). Target-body data were thin on some bodies (8 successful episodes in total on \ctrl{0.75}{0.45}, and 8 on task 2 for \ctrl{0.67}{0.3}), and in four of the six evaluated bodies the chosen $w=1$ is at the edge of the grid.

\textbf{The two stopping rules (added after bodies 1--5).}
They use calibration-task data only, so no transfer-task result was seen before stopping: stop if unguided success during collection is below 25\%, or if the best guided calibration-task success is less than 5 points above unguided. Both bodies were stopped by hand.
On the six evaluated bodies, the rules' inputs track the transfer-task gain (Pearson $r=0.996$), but the thresholds were set on five of them, all Panda controllers.
The second rule would also have stopped both no-room bodies and so hidden the significant drop.

\subsection{Robustness: seeds, evaluation seeds and episode limits}
\label{app:robust}
\label{app:kuka}
\label{app:ctrl-details}
\label{app:provenance}
\textbf{KUKA arm (25\% wrist).} The main number is the mean over three classifier seeds: 57\% (114/200), 52\% (104/200) and 48.5\% (97/200), against 42\% (84/200) unguided, all at $w=4$ (chosen for seed 0). Seeds 1 and 2 alone are not significant (+10 [0, +20], $p=0.064$; +6.5 [$-3$, +16], $p=0.21$).
Seed 0 raised success in three paired evaluations: 57\% vs.\ 42\%, 60\% (120/200) vs.\ 44.5\% (89/200) on a second evaluation seed, and 67.5\% (135/200) vs.\ 50.5\% (101/200) on 50 new object placements per task (+17 [+7, +26.5], $p=0.001$); pooled, 61.5\% (369/600) vs.\ 45.7\% (274/600).
The gain comes from more grasps, not from speed: failures that reach the object without grasping it fall from 57 to 35 episodes. Tasks 0, 6 and 7 gain in every evaluation, and task 3 loses in every one (pooled $-18.7$ [$-30.7$, $-7.3$] points; App.~\ref{app:failures}).
A classifier that also saw the gripper-to-object offset and the object lift gave no gain (44.5\%, 89/200).

\textbf{\ctrl{0.5}{0.3}, three classifier seeds (registered).} Two more seeds on the same data, with $w$ re-selected by the original rule, give 93\% (186/200), 93.5\% (187/200) and 91\% (182/200) against 73\% (146/200) unguided (pooled +19.5 [+13.5, +26]); the criterion (mean gain of at least 7.5 points, pooled CI and every seed's CI above zero) holds. The seeds share one data set, and seed 2 at $w=4$ hit the cap on 81\% of denoising steps (seed 1 at $w=1$: 22\%), so this does not test robustness to $w$.

\textbf{Episode limits.} On \ctrl{1}{0.2} at 560 steps, guidance raises success from 74.8\% (299/400) to 86.5\% (346/400) over two evaluation seeds (+11.75 [+6.75, +16.75]); read off the same episodes at 280 steps, the gain is 22.5 points, and guided episodes succeed sooner (mean time to success 256 vs.\ 324 steps). A speed-only re-timing rule (gain 1.25) scores 47.3\% (189/400), while sending the action planned 4 steps ahead scores 97.5\% (195/200) at 280 steps, equal to the source body's own unguided score. On \ctrl{0.5}{0.2}, guidance gives 95\% (190/200) vs.\ 73.5\% (147/200) at 280 steps, where every failure is a time-out, but only +3 [$-1$, +7] points at 560 steps: on this body, guidance mainly makes the robot finish sooner.

\textbf{Provenance.} Runs used two machines with the same code and seeds; for the second sibling-reuse pair and the cross-arm tests, the reference runs ran on the other machine one day earlier. Unfinished replicates are not reported.

\subsection{KUKA arm (25\% wrist): one round of guided collection}
\label{app:guided}
Guided collection ran guided \pz{} with the seed-0 classifier ($w=4$) on calibration tasks 1, 2, 4 and 5 (starting states 20--41, 22 episodes per task, 6,709 queries).
Guided \pz{} solved task 2 in 14 episodes, where the original target-body data had none.
We added the guided successes to the original target-body successes (10,772 chunks), retrained the classifier with the same recipe (seeds 0, 1 and 2) and chose $w$ again ($w=4$, 4 and 2).
Success is 73.5\% (147/200), 70\% (140/200) and 68.5\% (137/200), mean 70.7\%, against 57\%, 52\% and 48.5\% for the original classifier and 42\% unguided; per seed, the paired gain over the original classifier is +16.5 [+9.5, +24], +18 [+10.5, +25.5] and +20 [+12, +27.5] points.
Task 3 rises to 71.3\% on average (original classifier 41.3\%; unguided 54\%). The target-body cost is about 1.7$\times$ the original collection.

\emph{Equal-budget control (registered).} With about the same number of target-body \pz{} queries (6,520 vs.\ 6,709), a classifier retrained on unguided data reaches 51.5\% (103/200), 59\% (118/200) and 63\% (126/200), mean 57.8\%. Per-seed gaps to guided collection are +22 [+14.5, +29.5], +11 [+3.5, +18.5] and +5.5 [$-1.5$, +12.5] points, meeting the registered criterion (CI above zero in 2 of 3 seeds and a mean gap of at least 5 points). Extra unguided data add 5.3 points over the original classifier, and guidance adds 12.8 more; at the same budget, guided collection gathered 2.8$\times$ more task-2 success data (14 vs.\ 5 episodes).

\emph{Without the task-2 data (registered).} Using guided successes from tasks 1, 4 and 5 only gives 66.5\% (3-seed mean), meeting its criterion ($\geq65.5\%$) by 1 point. The task-2 data add about 4 points (pooled CI touches zero) and about half of the task-3 gain (task-3 means 41.3\%, 56\% and 71.3\% for the original classifier, this ablation and the full round).

\emph{A second round (registered).} The round-1 classifier collected 88 more guided episodes; an equal-episode control collected the same episodes with the original classifier. The second round gives 76\% (152/200), 78.5\% (157/200) and 71\% (142/200), mean 75.2\%, and the control 79\% (158/200), 70\% (140/200) and 74.5\% (149/200), mean 74.5\%; per seed, $-3$ [$-9.5$, +3.5], +8.5 [+1.5, +15.5] and $-3.5$ [$-10.5$, +3.5] points. A second round adds no detectable gain, although this test cannot rule out an effect of about 7.5 points either way.

\emph{Fresh scenes.} On 200 unused scenes, the seed-0 guided-collection classifier (the best of the three) scores 92\% (184/200) vs.\ 52\% (104/200) unguided (+40 [+32, +47.5]). No original-classifier or control condition ran on those scenes.

\emph{Caveats.} All three seeds share one guided data set; one evaluation seed; task 3 shaped the design; calibration task 1 does not gain (64\%, 66\% and 74\% vs.\ 68\% unguided, of 50). Guided successes are partly separable from unguided ones (probe AUC 0.72), and a registered drift flag was raised (source-body task-2 logit +2.15 at $t=0.5$, limit +2.0), so we do not claim that coverage alone explains the gain. Weaker steering alone does not: the original seed-0 classifier at $w=2$, which steers about as weakly, scores 54.5\% (109/200).

\emph{Other bodies.} On LIBERO-Goal, guided collection did not help \ctrl{0.5}{0.3} (67\% vs.\ 67\%, 134/200 each). On the Sawyer arm, guided \pz{} solved no task-2 episodes, so the round did not run.

\subsection{Other suites and second runs}
\label{app:suites}
All rows use one classifier and one evaluation seed, with $w$ chosen on calibration tasks 0, 2, 4 and 6.
Meeting the criterion: Spatial \ctrl{0.5}{0.3}, Spatial KUKA arm (+14 [+6.5, +21.5], $p=5\times10^{-4}$) and Goal \ctrl{0.33}{0.3} (+12 [+5.5, +19], $p=9\times10^{-4}$).
Inconclusive: Spatial \ctrl{1}{0.2} (+9 [0, +18], $p=0.06$; $w=4$ at the grid edge), Goal KUKA arm (+7 [0, +14], $p=0.07$; its classifier saw target-body data from tasks 2, 4 and 6 only, since the target body solved 0\% (0/21) of task-0 episodes), Goal \ctrl{1}{0.2} (+6.5 [$-0.5$, +13.5], $p=0.09$) and Goal \ctrl{0.5}{0.3} (+6.5 [$-0.5$, +14]).
Spatial \ctrl{0.33}{0.3} and \ctrl{0.5}{0.2} also have CIs above zero (Table~\ref{tab:main}).
No clear gain: all three LIBERO-10 bodies (at 980 steps, fixed before these runs because of LIBERO's 1,000-step cap) and the Kinova arm on Spatial and Goal. Two of these five rows start at 78\% unguided success or higher (LIBERO-10 \ctrl{0.5}{0.3}, Spatial Kinova), and on Goal the Kinova arm cannot do a calibration task (0\% (0/50) on calibration task 0); the LIBERO-10 KUKA and Kinova rows start near 40\% and have no simple explanation.

\emph{Second runs} repeat a run with fresh target-body collection seeds, a new classifier and a new $w$, keeping the source-body data, classifier seed, evaluation scenes and evaluation seed. Spatial \ctrl{0.5}{0.3}: +14.5 [+6, +22.5] (first run +17); Spatial KUKA arm: +12 [+4.5, +20] (first +14); Goal \ctrl{0.33}{0.3}: +7 [+1, +13.5] (first +12); Goal KUKA arm: +13.5 [+7, +20.5] (first +7); Goal \ctrl{0.5}{0.3}: +12.5 [+5.5, +19.5] (first +6.5). The last two were registered after their first runs came out inconclusive.

\emph{Second evaluation seeds.} With the same classifier and two evaluation seeds (criterion fixed before seed 1: pooled gain of at least 5 points with a CI above zero): Spatial \ctrl{0.5}{0.3} +18.5 [+12.5, +24.5], Spatial \ctrl{0.33}{0.3} +14 [+8.75, +19.25], Spatial KUKA arm +13.75 [+8.25, +19.25], Goal KUKA arm +8.75 [+3.75, +13.75] and Goal \ctrl{0.33}{0.3} +8 [+3.25, +12.75]; Spatial \ctrl{1}{0.2} +5.25 [$-1$, +11.5] is inconclusive.
One further run, the worn Panda on Spatial, drops from 93\% (186/200) to 81\% (162/200), $-12$ [$-18.5$, $-5.5$]; with unguided \pz{} already at 93\%, the registered rule says this body has no room, but the drop is significant.

\emph{Caveats.} The simulator is not deterministic: unguided runs on identical scenes and seed agree on only 68\% (136/200), 78.5\% (157/200) and 79.5\% (159/200) of episodes across the first three second-run pairs. Calibration task 0 drops under guidance in 6 of 8 new rows (e.g., Spatial \ctrl{0.33}{0.3}, 88\% to 52\%, of 50). Beyond the Holm correction over the 12 controller pairs (App.~\ref{app:stats}), the Goal \ctrl{0.33}{0.3} second run ($p=0.044$) would not survive a correction over all rows (Spatial \ctrl{0.5}{0.2}, $p=0.035$, already fails Holm); the Goal KUKA-arm and \ctrl{0.5}{0.3} second runs would ($p=3\times10^{-4}$ and 0.001).

\subsection{Reusing a classifier}
\label{app:reuse}
\label{app:xarm}
\label{app:shift}

\begin{table}[ht]
\centering
\scriptsize
\setlength{\tabcolsep}{2.2pt}
\begin{tabular}{@{}lllrrrll@{}}
\toprule
Suite & Target (no data) & Classifier from & Ung. & Reused & Own & Reused $-$ ung. & Reused $-$ own \\
\midrule
Object & \ctrl{0.5}{0.3} & \ctrl{0.5}{0.2} & 73 (146) & 96 (192) & 93 (186) & +23 [+16.5, +29.5] & +3 [$-$1.5, +7.5] \\
Object & \ctrl{0.5}{0.2} & \ctrl{0.5}{0.3} & 94.5 (189) & 97 (194) & 97.5 (195) & +2.5 [$-$1.5, +6.5] & $-$0.5 [$-$4, +3] \\
Goal & \ctrl{0.33}{0.3} & \ctrl{0.5}{0.3} & 72.3 (144.5)$^a$ & 82 (164) & 81.8 (163.5)$^b$ & +9.75 [+3.75, +15.5] & +0.25 [$-$5.75, +6] \\
\midrule
Object & Kinova (10\% wrist) & KUKA (25\% wrist) & 70 (140) & 58.5 (117) & 85 (170) & $-$11.5 [$-$19.5, $-$4] & $-$26.5 [$-$34.5, $-$18.5] \\
Object & Kinova (10\% wrist) & KUKA, guided coll. & 70 (140) & 65.5 (131) & 85 (170) & $-$4.5 [$-$13.5, +4.5] & $-$19.5 [$-$27.5, $-$11.5] \\
Object & Panda (10\% wrist) & KUKA (25\% wrist) & 47 (94) & 24.5 (49) & 37.5 (75) & $-$22.5 [$-$31, $-$14] & $-$13 [$-$21, $-$5] \\
\bottomrule
\end{tabular}
\caption{\textbf{Reusing a classifier on a body that contributed no data.} Success in \% with successes out of 200 in parentheses; differences in points. Top: sibling controller bodies (the second row is at the ceiling). Bottom: across arms. Own: the target's own classifier; guided coll.: the KUKA arm's guided-collection classifier. $^a$Mean of unguided runs 69.5\% (139/200) and 75\% (150/200). $^b$Mean of its two classifiers, 81.5\% (163/200) and 82\% (164/200).}
\label{tab:reuse}
\end{table}

\emph{Sibling bodies (registered).} The target uses, unchanged, the embodiment classifier and $w$ ($w=1$) of its nearest sibling by stiffness and delay; it contributes no data.
On LIBERO-Object, the classifier of \ctrl{0.5}{0.2} lifts \ctrl{0.5}{0.3} by 23 points, with no difference shown from the target's own classifier (one pair; the two conditions ran two days apart). In the reverse direction the target is at the ceiling, so the registered reuse criterion could not be met; a ``matches own'' check fixed before data held, and in 89\% (178/200) of pairs all three conditions succeed.
On LIBERO-Goal, the second-run classifier of \ctrl{0.5}{0.3} lifts \ctrl{0.33}{0.3} by 9.75 points over the registered reference (the mean of two unguided runs; +12.5 and +7 against each alone), matching the target's own classifiers.

\emph{Across arms (registered).} Each test used a KUKA-arm classifier at the KUKA arm's own $w=4$, with no target data. None met its reuse criterion, and two drops (the original classifier on the Kinova arm and on the Panda, $p=1.4\times10^{-6}$) survive a Bonferroni correction over the four reuse tests. Classifier and $w$ changed together (at $w=4$ the cap binds on 64--74\% of denoising steps on the Kinova arm and 49--54\% on the Panda, against 9--23\% for their own classifiers at $w=1$), so these runs show no cause.

\emph{Shared shift (registered prediction).} On saved successful calibration-task chunks, a linear probe (5 folds by episode) cannot tell the siblings \ctrl{0.5}{0.2} and \ctrl{0.5}{0.3} apart (AUC 0.51 at $t=0$) but separates the KUKA and Sawyer arms, both with 25\% wrists, almost perfectly (0.99). The siblings' guidance gradients align (cosine +0.71 at $t=0.3$; +0.57 to +0.60 across other seeds) while the arms' do not (+0.07). Both registered gaps hold (AUC gap 0.481, bar 0.10; cosine gap 0.632, bar 0.20).
Caveats: the Sawyer arm stands in for the Kinova arm (no Kinova chunks were saved); both sibling shifts are small (probe AUC against the source body 0.62 and 0.66, against 0.99 and 0.98 for the arms); shared seed-0 initialization inflates the sibling cosine. Because chunks carry the end-effector pose (App.~\ref{app:setup}), part of the arms' separability may reflect different poses rather than different motions; we did not test this.

\subsection{Baselines}
\label{app:baselines}

\begin{table}[ht]
\centering
\scriptsize
\setlength{\tabcolsep}{3.5pt}
\begin{adjustbox}{max width=\textwidth}
\begin{tabular}{@{}llr>{\raggedright\arraybackslash}p{1.75in}@{}}
\toprule
Body & Condition & Success, \% (n/200) & Note (points) \\
\midrule
KUKA arm (25\% wrist) & unguided & 42 (84) & \\
 & LoRA, same query budget & 29.5 (59) & $-$12.5 [$-$21, $-$4]; 43 of 56 training successes from one task \\
 & LoRA, re-collected successes & 41.5 (83) & $-$0.5 [$-$10, +9] \\
 & Best-of-$N$ ($N=8$) & 35.5 (71) & $-$6.5 [$-$15.5, +2]; no gradient \\
 & guided (guided-collection clf., seed 0) & 73.5 (147) & vs.\ Best-of-$N$ +38 [+29.5, +46]; vs.\ LoRA +32 [+24, +40] \\
 & rotation command $\times$2 & 21 (42) & $-$21 [$-$28.5, $-$13.5] \\
 & guided (seed 0) / grasp-gated & 57 / 55 (114 / 110) & gated vs.\ ungated $-$2 [$-$11, +7] \\
Panda (locked wrist) & unguided / guided / gated & 68.5 / 36 / 67 (137 / 72 / 134) & gating restores the unguided level \\
Panda (slow gripper) & unguided / guided / gated & 76 / 73.5 / 78.5 (152 / 147 / 157) & no difference significant \\
UR5e (speed cap) & unguided / guided / rule / oracle & 84 / 77.5 / 85.5 / 52.5 (168 / 155 / 171 / 105) & rule vs.\ guided +8 [+1, +15] \\
\ctrl{0.5}{0.3} & unguided / Best-of-$N$ / guided & 73 / 76 / 93 (146 / 152 / 186) & $N=4$: +3 [$-$4, +10]; guided vs.\ it +17 [+10.5, +23.5] \\
\bottomrule
\end{tabular}
\end{adjustbox}
\caption{\textbf{Baselines and the grasp-triggered gate} (LIBERO-Object, 560 steps). Notes give the difference to unguided \pz{} in points unless stated. Best-of-$N$: the embodiment classifier used for guidance picks the highest-logit chunk among $N$ unguided samples (KUKA arm: the seed-0 guided-collection classifier, compared with gradient guidance using that same classifier; \ctrl{0.5}{0.3}: its own seed-0 classifier). The gate switches guidance on at the first executed gripper close; it was designed from failures on KUKA task 3. Rule: a hand-made UR5e rule (rotation gain 0.3, smoothing 0.6); oracle: a state-aware speed limiter.}
\label{tab:gate}
\end{table}

\begin{itemize}
\item \textbf{Best-of-$N$ (registered, two bodies).} For each query, \pz{} samples $N$ chunks without guidance, and the robot runs the one with the highest classifier logit. $N$ was chosen from $\{4,8,16\}$ on calibration tasks (KUKA arm: 21.3\% (17/80), 25\% (20/80) and 22.5\% (18/80), so $N=8$; \ctrl{0.5}{0.3}: 80\% (64/80), 78.8\% (63/80) and 75\% (60/80), so $N=4$). Within one state, \pz{}'s samples differ by a median of only 0.1--0.2 (KUKA) and 0.2--0.5 (\ctrl{0.5}{0.3}) logits. A Best-of-$N$ query took 126--370\,ms and a guided query 134--142\,ms under different GPU loads, so we make no equal-compute claim.
\item \textbf{LoRA (KUKA arm).} A fixed, untuned recipe: LoRA on \pz{}'s action expert, 2,000 steps, batch 16, learning rate $10^{-4}$, no replay, one seed. ``Same budget'' trains on 56 successful episodes from an unguided collection of the original budget. ``Re-collected'' trains on target-body successes collected with the same protocol as the guided-collection classifier's data (26\% larger; exact replay reproduced only 89.9\% of steps, below a 95\% gate, so the data were re-collected). We do not claim that LoRA cannot adapt.
\item \textbf{Hand-made rules.} On the UR5e, $w$ and the rule were selected at 280 steps and not re-selected at 560. On \ctrl{1}{0.2}, see App.~\ref{app:robust}.
\item \textbf{Not run.} We did not compare against a trained residual model such as J-PARC~\cite{jo2026jparc}, or against DSRL~\cite{wagenmaker2025dsrl}, which needs a reward and online episodes per task.
\end{itemize}

\section{Failure Analysis}
\label{app:failures}
Analyses without a registered criterion are labeled descriptive.

\textbf{Panda (locked wrist).} Guidance ($w=2$) gives 36\% (72/200) vs.\ 68.5\% (137/200) unguided, and every transfer task drops. Many guided plans ask for wrist roll that the locked joint cannot make. Switching guidance on only at the first executed gripper close restores the unguided level (67\%, 134/200; $-1.5$ [$-9.5$, +7] points), so the harm comes mostly from guidance during the approach.

\textbf{Sawyer arm (25\% wrist).} 25\% (50/200) vs.\ 41\% (82/200) unguided ($-16$ [$-23$, $-9$] points). Its embodiment classifier saw target-body data from calibration tasks 1, 4 and 5 only (task 2: no successes). Calibration task 1 rises (60\% to 72\%, of 50) while every transfer task falls, which points to a coverage gap.

\textbf{Panda (10\% wrist).} 37.5\% (75/200) vs.\ 47\% (94/200) ($-9.5$ [$-17$, $-2$] points). Transfer task 6 is at 0\% (0/50) unguided, and calibration task 1 also drops (60\% to 42\%). On the two calibration tasks without target-body successes (2 and 4), guided \pz{} succeeded on at most 10\% (2/20) of episodes per task and $w$, so guided collection would add almost no data.

\textbf{KUKA arm, task 3 (descriptive).} Task 3 is a tall bottle in the center slot; the other transfer tasks use side slots, and the target body had no successes on task 2, the calibration task with a center-slot tall object. Under guidance, failures in which the hand never reaches the object rise (13 to 18 and 12 to 17 episodes on two evaluation seeds; 2 to 14 on new scenes), and task-3 approach chunks get the lowest classifier logit and the largest gradient. The grasp gate moves the harm from task 3 to task 6, and guided collection, which adds task-2 data, lifts task 3 (App.~\ref{app:guided}).

\textbf{UR5e (speed cap).} 77.5\% (155/200) vs.\ 84\% (168/200) at 560 steps ($-6.5$ [$-13$, 0] points; McNemar $p=0.08$). A hand-made rule (rotation gain 0.3, smoothing 0.6) beats guidance by 8 [+1, +15] points, and a state-aware oracle speed limiter is the worst condition (52.5\%, 105/200). Descriptively, the UR5e's successful calibration-task chunks command about 17\% less rotation than the source body's and the same translation, yet 69--82\% of the classifier gradient's energy lies on the translation dimensions, so guidance probably redirects translation rather than trimming rotation (guided actions were not logged).

\textbf{Panda (slow gripper) and other near misses.} With a slow gripper, guidance gives 73.5\% (147/200) vs.\ 76\% (152/200); a slow gripper changes only the timing of the close, which chunk-shape guidance hardly addresses. The worn Panda shows no significant change (48\% vs.\ 43\%). \ctrl{0.33}{0.3} on LIBERO-Object gains 9 [+4, +14] points ($p=0.0009$) but misses its registered criterion of 10 points, with unguided \pz{} near the ceiling (88.5\%).

\textbf{Panda (locked upper arm).} Its unguided pilot (14\%) was outside the 30--75\% band, so it is not a body-level result. Tasks 0, 6 and 7 fail in both conditions; on task 3, guidance gives 84\% (42/50) vs.\ 44\% (22/50) (paired $p=9\times10^{-5}$).

\textbf{Predicting where guidance helps.} None of our registered predictors worked: body separability vs.\ gain (Spearman $-0.14$ on 6 bodies; bar 0.70), a failure-predicted gate (loses 4--7.5 points to always guiding on four KUKA-arm conditions), a confidence gate (0 of 4 conditions), and an open-loop replay forecast of unguided failure (within-body Spearman $-0.18$; bar 0.4). Descriptively, on the KUKA arm, guidance helps where early chunks already look target-body-like (logit AUC 0.65--0.71) and hurts where the classifier gradient is largest (AUC 0.66--0.73).

\textbf{Cross-suite transfer (registered, failed).} A LIBERO-Object embodiment classifier for \ctrl{0.5}{0.3} does not transfer to LIBERO-Goal: AUC 0.55 at $t=0.1$ and 0.61 at $t=0.5$ (bars 0.85 and 0.80), and its mean guidance direction has cosine $-0.22$ and $-0.36$ with the Object-suite direction (bar 0.7). Goal-suite chunks sit far from Object-suite chunks in \pz{}'s action space (per-dimension means differ by up to about one standard deviation, because that space subtracts the end-effector pose), so part of the failure may be a layout shift rather than a body difference. Each suite therefore needs its own target-body data.

\section{Extended Related Work}
\label{app:related}
\textbf{Guidance with a domain classifier.}
Classifier guidance adds a noised-sample classifier's gradient to a frozen model's score~\cite{dhariwal2021diffusion}; classifier-free guidance~\cite{ho2022cfg} would need \pz{} retrained with a body label.
Our operator is the density-ratio correction of discriminator guidance~\cite{kim2023discriminator} and TGDP~\cite{ouyang2024tgdp}, which add a classifier-derived density-ratio gradient (real vs.\ generated, or target vs.\ source; TGDP through a learned guidance network) to a frozen score model. X-Diffusion~\cite{pace2025xdiffusion} uses a noise-level human-vs-robot classifier to decide from which noise level each human action enters training, and DARC~\cite{eysenbach2021darc} corrects RL rewards for dynamics gaps with domain classifiers.

\textbf{Steering frozen robot policies.}
Most steering signals encode the task or its success: a success predictor~\cite{wang2026ppguide}, a value function~\cite{nakamoto2024vgps}, human input~\cite{wang2025itps}, a dynamics model~\cite{du2025dynaguide}, a latent barrier~\cite{sun2025lpb}, proxy policies~\cite{ning2026pps}, guided-rollout retraining~\cite{lin2026reguide} or a noise-space actor~\cite{wagenmaker2025dsrl}.
Real-time chunking guides a flow VLA's sampler under inference delay~\cite{black2025rtc}, and UMI-on-Air~\cite{gupta2025umionair} steers a diffusion policy with a known controller's tracking cost; we learn the correction from success-labeled rollouts.

\textbf{Embodiment transfer.}
Prior methods fine-tune on target demonstrations~\cite{zhang2025ate}, bridge appearance rather than dynamics~\cite{chen2024mirage,lepert2024shadow,chen2024roviaug}, edit weights~\cite{kang2026dart}, condition on joint health~\cite{arslan2026health}, or train a residual calibrator for joint faults from paired rollouts~\cite{jo2026jparc}. We need no joint data or paired rollouts, but do not compare to them.

\end{document}